\documentclass[conference]{IEEEtran}
\IEEEoverridecommandlockouts
\usepackage[letterpaper, left=1.01in, right=1.1in, bottom=1.1in, top=0.75in]{geometry}
\usepackage{cite}
\usepackage{amsmath,amssymb,amsfonts}
\usepackage{algorithmic}
\usepackage{graphicx}
\usepackage{textcomp}
\usepackage{orcidlink} 
\usepackage{subcaption}
\usepackage{multirow}
\usepackage{array}
\usepackage{listings}
\usepackage{tabularx}
\usepackage{indentfirst}
\usepackage{color}
\usepackage{hyperref}
\usepackage{tabularray}
\usepackage{comment}

\makeatletter
\def\@citex[#1]#2{\leavevmode
\let\@citea\@empty
\@cite{\@for\@citeb:=#2\do
{\@citea\def\@citea{,\penalty\@m\ }%
\edef\@citeb{\expandafter\@firstofone\@citeb\@empty}%
\if@filesw\immediate\write\@auxout{\string\citation{\@citeb}}\fi
\@ifundefined{b@\@citeb}{\hbox{\reset@font\bfseries ?}%
\G@refundefinedtrue
\@latex@warning
{Citation `\@citeb' on page \thepage \space undefined}}%
{\@cite@ofmt{\csname b@\@citeb\endcsname}}}}{#1}}
\makeatother
\def\BibTeX{{\rm B\kern-.05em{\sc i\kern-.025em b}\kern-.08em
    T\kern-.1667em\lower.7ex\hbox{E}\kern-.125emX}}

\ifCLASSOPTIONcompsoc
  \usepackage[nocompress]{cite}
\else
  \usepackage{cite}
\fi
\ifCLASSINFOpdf
\else
\fi

\DeclareRobustCommand*{\IEEEauthorrefmark}[1]{%
  \raisebox{0pt}[0pt][0pt]{\textsuperscript{\footnotesize #1}}%
}

\begin{document}
\title{Ensemble Learning for Vietnamese Scene Text Spotting in Urban Environments}

% \author[1,2]{Hieu T. Nguyen \orcidlink{0009-0001-7726-5301}$^\dagger$\thanks{$\dagger$Equal contribution}}
% \author[1,2]{Cong-Hoang Ta\orcidlink{0009-0009-7946-736X}$^\dagger$}
% \author[1,2]{Phuong-Thuy Le-Nguyen\orcidlink{0009-0005-0493-581X}$^\dagger$}
% \author[1,2]{Trung-Nghia Le\orcidlink{0000-0002-7363-2610}}
% \affil[1]{Faculty of Information Technology, University of Science, VNU-HCM, Vietnam}
% \affil[2]{Vietnam National University, Ho Chi Minh City, Vietnam}

\author{%
\IEEEauthorblockN{%
Hieu Nguyen\orcidlink{0009-0001-7726-5301}%\IEEEauthorrefmark{1,}\IEEEauthorrefmark{2}
\textsuperscript{\rm 1,2,$\dagger$}\thanks{$\dagger$Equal contributors}, Cong-Hoang Ta\orcidlink{0009-0009-7946-736X}%\IEEEauthorrefmark{1,}\IEEEauthorrefmark{2}
\textsuperscript{\rm 1,2,$\dagger$},
Phuong-Thuy Le-Nguyen\orcidlink{0009-0005-0493-581X}%\IEEEauthorrefmark{1,}\IEEEauthorrefmark{2}
\textsuperscript{\rm 1,2,$\dagger$}, \\
Minh-Triet Tran\orcidlink{0000-0003-3046-3041}\textsuperscript{\rm 1,2}%\IEEEauthorrefmark{1,}\IEEEauthorrefmark{2}
, Trung-Nghia Le\orcidlink{0000-0002-7363-2610}%\IEEEauthorrefmark{1,}\IEEEauthorrefmark{2}
\textsuperscript{\rm 1,2,*}\thanks{*Corresponding author. Email address: ltnghia@fit.hcmus.edu.vn}\\}%

\IEEEauthorblockA{%
\IEEEauthorrefmark{1}%
\textit{Faculty of Information Technology, University of Science, VNU-HCM, Vietnam}\\%
\IEEEauthorrefmark{2}%
\textit{Vietnam National University, Ho Chi Minh City, Vietnam}\\%
\ }
}
\maketitle
\begin{abstract}
This paper presents a simple yet efficient ensemble learning framework for Vietnamese scene text spotting. Leveraging the power of ensemble learning, which combines multiple models to yield more accurate predictions, our approach aims to significantly enhance the performance of scene text spotting in challenging urban settings. Through experimental evaluations on the VinText dataset, our proposed method achieves a significant improvement in accuracy compared to existing methods with an impressive accuracy of 5\%. These results unequivocally demonstrate the efficacy of ensemble learning in the context of Vietnamese scene text spotting in urban environments, highlighting its potential for real-world applications, such as text detection and recognition in urban signage, advertisements, and various text-rich urban scenes. 
\end{abstract}


%\highlight{CHU Y: NHUNG GI SUA LAI THI DE VAO TRONG HIGHLIGHT DE DE THEO DOI. THAY KHONG CO THOI GIAN DE DOC HET CA PAPER.}

\section{INTRODUCTION}
 The detection and recognition of text in images, namely \textit{scene text spotting}, represent a highly formidable task in computer vision\cite{huang2022swintextspotter}, demanding the precise localization and identification of text sequences within real-world contexts\cite{liu2023spts}. The implications of scene text spotting are far-reaching, spanning crucial applications such as key entities recognition \cite{wang2021robust}, autonomous driving \cite{9551780}, intelligent navigation \cite{rong2016guided,7353895}, etc. Scene text spotting has been recently drawing much attention from the community, resulting in rapid investigations \cite{9989248,10.1007/978-3-031-15063-0_34}.%\highlight{them vai ref nua (DA THEM)}. 
 %CNNs have proven to be successful in various computer vision tasks, including scene text spotting, image recognition, and object detection\cite{8308186}. %, which aims at simultaneous text detection and recognition in natural images. 

 Nonetheless, the intricacies of the Vietnamese language, with its rich set of characters and diacritics, pose significant challenges for scene text spotting. Certain Vietnamese characters, especially when accompanied by diacritics, exhibit visual similarities that can lead to confusion, such as, `ă' and `â', `ô' versus `ó', `é' and `ê'. Consequently, conventional individual models may encounter limitations when it comes to accurately detecting and recognizing Vietnamese scene text.
 %\textcolor{blue}{(See Fig. \ref{urbansceneslabel}). }

 % . CNNs have been widely used for 
% \COMMENT{passive..}
 
 In addition, the realm of scene text spotting in Vietnamese urban environments is rife with challenges for existing methods \cite{tran2022improving,huang2022swintextspotter}
 %\highlight{ref cua cac pp co kq thap tren tap vintext (DA THEM)}. 
 In bustling city contexts, scene texts are frequently occluded, either partially or entirely, due to the presence of diverse surrounding objects such as trees, towers, traffic signs, electric poles, etc. The variability in lighting conditions, encompassing weather fluctuations and low contrast scenarios caused by background reflections, further exacerbates the difficulty of scene text spotting.

 \begin{figure}[t]
    %\centering
    \includegraphics[width=\linewidth]{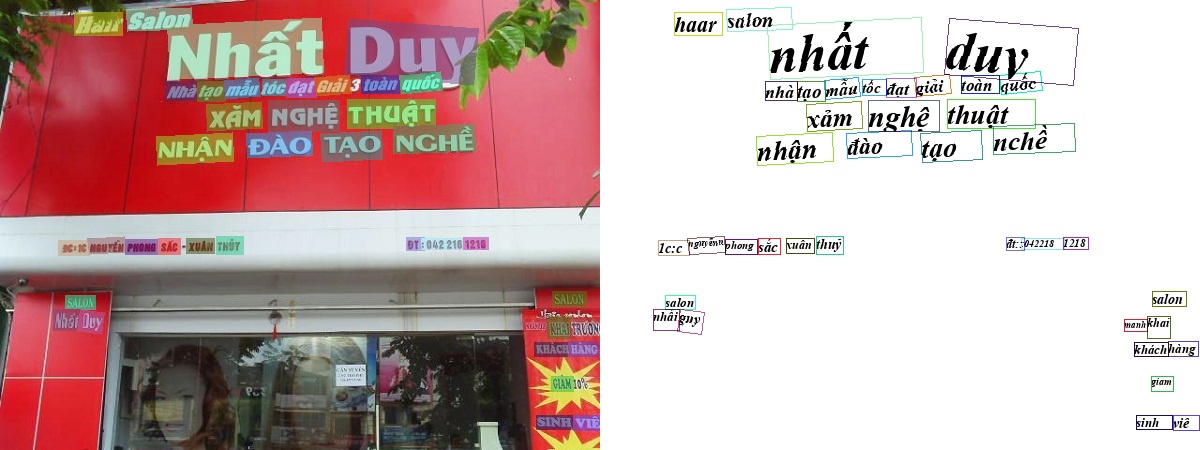}
    %\centerline{\includegraphics[height=5cm]{urban_scenes.png}}
    \newline
    
    \centering
    \includegraphics[width=\linewidth]{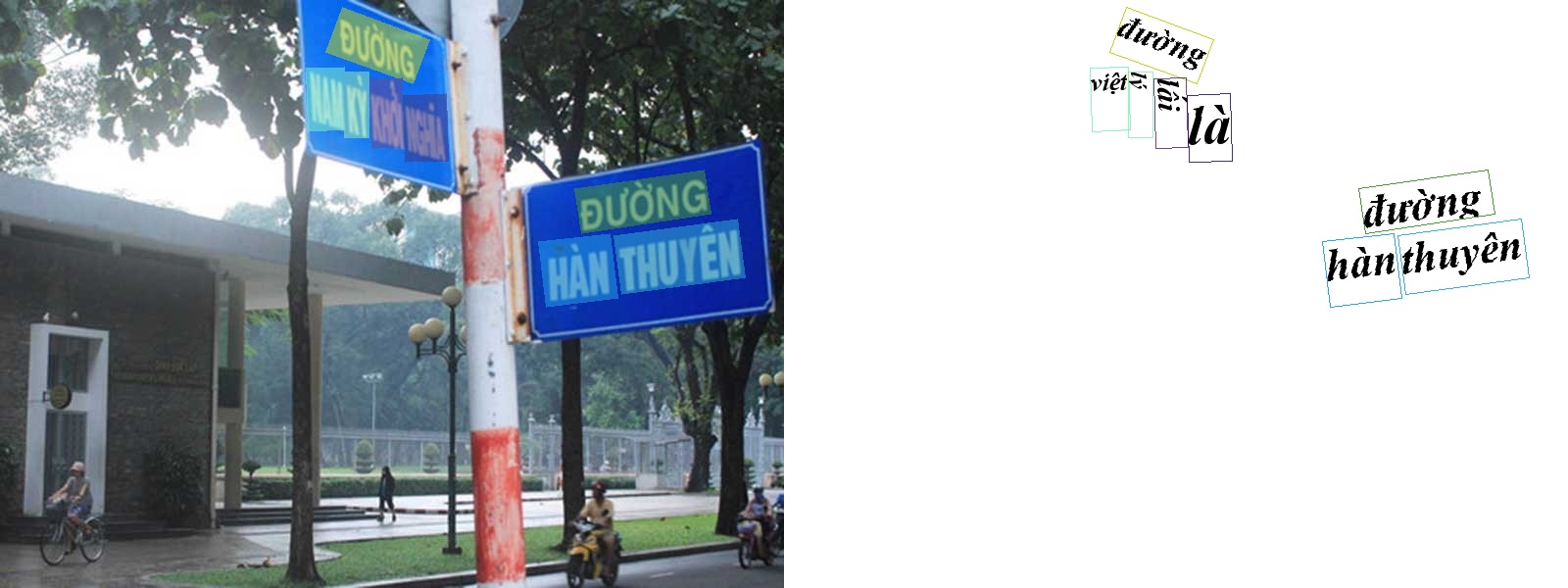}
    % \captionsetup{justification=centering}
    \caption{Scene text spotting in Vietnamese urban environments poses various challenges, such as obscured by trees and perspective-shifted.}
    \label{fig:urbansceneslabel}
\vspace{-5mm}
\end{figure}

% \COMMENT{tách ý của đoạn 2 thành đoạn riêng, vd sts đối mặt vs j một cách chung chugn, nêu thêm 1 số vd về vietnamese}
%\textcolor{red}{doan nay thay bao viet cu the chu khong viet motivate by limitation kieu vay ay'}

 Existing approaches predominantly rely on individual models~\cite{nguyen2021dictionary,le2019end,liu2023spts,9989248},
 %\highlight{ref (DA THEM)}, 
 each exhibiting varying degrees of success across different subsets of images. However, when faced with the complexities of Vietnamese urban environments, an intricate domain with large and diverse contexts, such single-model solutions often yield limited results. 
 
 To overcome these limitations, we propose an ensemble learning framework, a powerful technique that has shown remarkable efficacy in various computer vision tasks\cite{yu2020accurate,fang2021read}. Our ensemble learning framework combines multiple state-of-the-art methods, leveraging their individual strengths and mitigating their weaknesses. Indeed, each method is tailored to address distinct challenges related to text detection, recognition, and the intricacies of the Vietnamese script. By aggregating their predictions, we can significantly enhance the performance of Vietnamese scene text spotting in urban environments. Extensive experiments on the VinText dataset \cite{nguyen2021dictionary} show that our proposed ensemble learning framework outperforms individual models for Vietnamese scene text spotting, demonstrating superior accuracy and robustness in challenging urban environments by boosting the accuracy up to 5\%. The experimental findings underscore the potential of ensemble learning as a powerful tool for advancing scene text spotting in dynamic urban environments.
 
%%%%%%%%%%%%%%%%

% In this paper, we realize a comprehensive ensemble learning framework specifically tailored for Vietnamese scene text spotting in urban environments. The proposed framework consists of multiple state-of-the-art models, each addressing specific challenges related to text detection, recognition, and the characteristics of the Vietnamese script. By fusing the outputs of these models, we aim to achieve improved accuracy and robustness in scene text spotting.

 %We propose an ensemble learning approach, leveraging the power of Convolutional Neural Networks (CNNs). 
 
 % \COMMENT{chi can 1 cau, viet ngan gon lai, kết quả cải tiến ghi tạm thời là 5 phần trăm nhé, phần abstract cũng thế }
 % To evaluate the effectiveness of our ensemble learning approach, we used a benchmark VinAI dataset comprising diverse urban scenes captured in various cities across Vietnam. We conducted extensive experiments and compared the performance of our ensemble framework against individual models and other existing methods designed for Vietnamese scene text spotting. The experimental results of up to 5\% demonstrate the superiority of our ensemble approach, showcasing enhanced accuracy and robustness in challenging urban environments.

 Our paper contributes to the field of Vietnamese scene text spotting in urban environments in the following ways:
 \begin{itemize}
 \item We propose a novel ensemble learning framework specifically designed for Vietnamese scene text spotting, addressing the complexities of urban environments and the unique characteristics of the Vietnamese script.
 \item We conduct thorough experiments using an extensive dataset encompassing diverse urban scenes in Vietnam. Widely adopted in the research community, this dataset enables a comprehensive evaluation of our proposed approach and facilitates fair comparisons with existing methods. Extensive experimental findings demonstrate the utility and superiority of our ensemble approach, exhibiting higher accuracy and enhanced robustness in challenging urban contexts.
 \end{itemize}
 
% In summary, our work strives to advance the state-of-the-art in Vietnamese scene text spotting and offers a valuable contribution to the field of computer vision, particularly in urban environments.
% <viết tên section theo format của bài và có ref đến phần dưới luôn nhé>
% The remainder of this paper is structured as follows: Section \ref{sec:related_work} reviews related work on scene text spotting. Section \ref{sec:proposed} describes our proposed ensemble learning framework for Vietnamese scene text spotting in urban environments in detail. Section \ref{sec:experiemnts} presents the experimental results that demonstrate the effectiveness of our method. Finally, Section \ref{sec:conclu} concludes our work and suggests directions for future research.

 \begin{figure*}[t]
    \vspace{1mm}
    \centering
    \includegraphics[width=\textwidth]{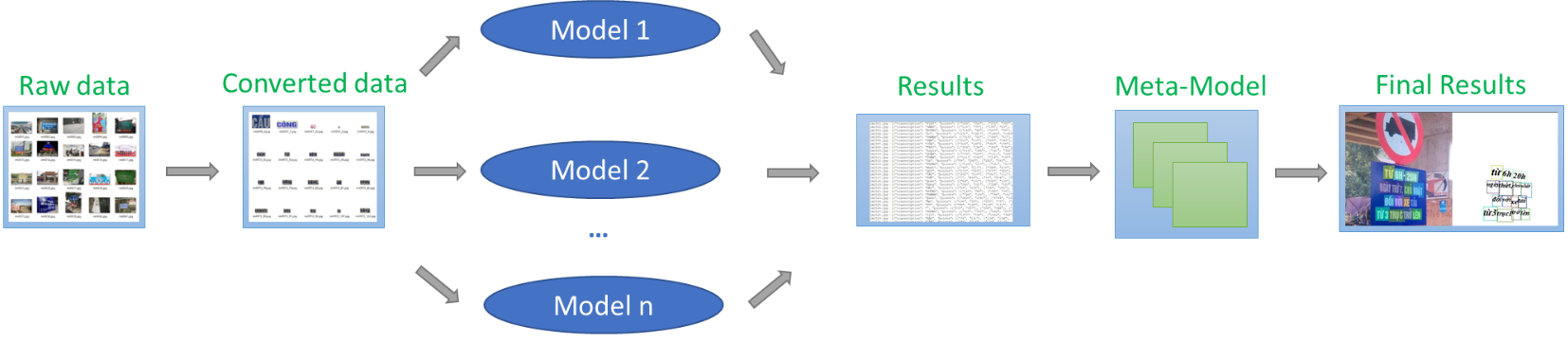}
    \caption{Workflow of the proposed ensemble learning framework for Vietnamese scene text spotting.}
    \label{ensemblewf}
%\vspace{-3mm}
\end{figure*}

\section{RELATED WORK}
\label{sec:related_work}

\subsection{Scene Text Spotting}

In recent decades, the growth of deep learning significantly contributes to the advancement of scene text spotting. There are typically approaches: two-stage and end-to-end approaches. The two-stage approach consists of two major stages: scene text detection and scene text recognition. A plethora of effective algorithms have been proposed for each stage, including DB/DB++ \cite{DB, DB++}, SAST \cite{SAST}, EAST \cite{EAST}, etc. for detection and SPIN \cite{SPIN}, SRN \cite{SRN}, SVTR \cite{SVTR}, ABINet \cite{ABINet}, etc. for recognization. Efforts have also been directed towards integrating detection and recognition processes, with works like TextBoxes by Liao et al. \cite{LiaoSBWL} combining a single-shot detector and a text recognizer. Furthermore, Nguyen et al. \cite{nguyen2021dictionary} leveraged a dictionary to generate a list of potential results and identify the most visually compatible outcome with the text's appearance to train and improve the recognition stage. Additionally, He et al. contributed VinText dataset \cite{nguyen2021dictionary}, a challenging benchmark for Vietnamese scene text spotting. Despite these advancements, existing techniques treat detection and recognition as independent tasks, lacking seamless information exchange between them.

% \subsection{End-to-End Scene Text Spotting}

%\highlight{trong section nay dung thi qua khu don cho nhung cau: ai do lam gi do. thay da sua giup phan dau. Tui em tu sua cac phan tiep theo}
%\highlight{TODO: CHECK LAI}
Meanwhile, end-to-end approach focuses on merging detection and recognition into a unified system. Peng et al. \cite{peng2022spts} presented an end-to-end scene text spotting method that approaches scene text spotting as a sequence prediction task. Wang et al. proposed PGNet \cite{PGNet}, which revolves around developing a model that combines a detection unit and a recognition module, allowing for shared CNN features and joint training. Huang et al. \cite{huang2022swintextspotter} introduced a novel end-to-end scene text spotting framework called SwinTextSpotter. By integrating both functionalities into a single algorithm, the resulting end-to-end model becomes more compact and efficient, leading to improved speed and performance
recognition. 
%\highlight{kiem tra ky doan nay, thay sai nhieu cho. kiem them mot vai method nua.}

\subsection{Ensemble Learning}
Ensemble learning represents a powerful methodology that amalgamates the strengths and benefits of multiple approaches, culminating in a superior model. As early as 1996, Rosen proposed \cite{rosen1996ensemble} a technique called decorrelation network training to enhance the accuracy of regression learning in ensemble neural networks. 
Deng et al. \cite{deng2014ensemble} introduced linear and log-linear stacking methods for ensemble learning, focusing on applications to speech class posterior probabilities computed by convolutional, recurrent, and fully-connected deep neural networks.
Recently, Casado-García and Heras \cite{ensembleObj} explored ensemble methods for object detection, addressing the challenges associated with ensembling object detectors, such as the limitations of existing ensemble approaches that depended on specific detection models or frameworks. 
%\highlight{revise doan nay. kiem them mot vai method nua.}
Leveraging the advantage of different methods, we introduce an ensemble learning framework to integrate results from individual models of both the two-stage and end-to-end approaches to improve the performance of scene text spotting. 

\section{PROPOSED METHOD}
\label{sec:proposed}
\subsection{Overview}

%\highlight{CHU Y cho TOAN BO paper: khong in dam neu khong thuc su can thiet}

%\begin{figure}[t]
%\centerline{\includegraphics[width=3.5in]{Figures/rec_format.png}}
%\caption{Data and labels format to train recognition model \cite{ocr}}
%\label{rec_format}
%\end{figure}

We present an ensemble learning framework designed to effectively combine the outputs of multiple scene text spotting methods. Fig. \ref{ensemblewf} illustrates the overview of our method, which consists of three main components: data converter, base models, and meta-model.

In the first stage of the workflow, we transform the raw images and labels of the VinText dataset \cite{nguyen2021dictionary} into different formats suitable for training and predicting of each base model. The data used to train the recognition model is images with a single word, while the dataset includes images with many different text lines which contain more than one word. Therefore, we crop the original images to smaller images that contain only one word and also create new labels for the new images. 

After that, base models are trained on converted data and used for prediction to obtain initial results. Subsequently, the results of all models are combined with an ensemble technique and are essential data to create a meta-model. The meta-model is then used for prediction and generates final result, which is expected to be better than the initial ones. 

%\textcolor{blue}{Nho viet lai doan nay}
%Two techniques are proposed to combine and create Meta-Model: Text Box Combination and ... are described in subsection B and C respectively

\subsection{Text Box Combination:}

%\highlight{viet doan nho gioi thieu cho phan nay. vi du input, output, muc dich.... Enhancing Recognition Results through Aggregation}

%\highlight{qua trinh gom 2 phan merge non-overlap va overlap?}
%\textcolor{red}{NOTE: Doi voi moi cap 2 anh khi merge se tuy xem overlap hay khong de merge theo 1 trong 2 cach}

%When combining text boxes from the results of two different models, we propose two merging approaches based on whether each pair of text boxes has an overlap or not: merging overlapping text boxes and merging non-overlapping text boxes.

Our posed Meta-model undergoes two processes: merging non-overlapping text boxes and merging overlapping text boxes. Given an image and $n$ base models, the input of our ensemble algorithm is a list of prediction results: $D = [D_1,\ldots, D_n]$ where each $D_i$, with $i\in\{1,\ldots,n\}$, is a list of predicted text boxes $D_i = [B_1, B_2,\ldots]$. In general, each $D_i$ comes from the predictions of a scene text spotting model using a particular method $M_i$ for a given image. 

\subsubsection{Merging non-overlapping text boxes}
\label{merge:1}
%Ensemble result $D_s$ is generated by grouping non-overlapping text boxes of $D_i$ together based on the following formula: 
Firstly, we combine the non-overlapping text boxes belonging to different $D_i$ results together to form the result $D_s$:

\begin{equation}
    D_s = D_1 \cup D_2 \cup \ldots \cup D_n, 
\end{equation}
where pair of non-overlapping text boxes ($B_h, B_k$) are defined by:
\begin{equation}
\begin{cases}
\begin{aligned}
%B_j \in D_i, B_k \notin D_i \\
\forall P_o \in B_h\ \implies P_o \notin B_k, \\
\forall P_o \in B_k\ \implies P_o \notin B_h, \\
%\forall P_o \in D_i\, ,\, \exists P_o \in B_j\ \implies P_o \notin B_k,
\end{aligned}
\end{cases}
\end{equation}
where $B_h, B_k$ are text boxes consisting of the coordinates of four points at the four corners of the text box, starting from the bottom left corner in counterclockwise order. $P_o$ is the coordinates of a point in Oxy, which is one of four points of a text box (B). 
%\highlight{chua dinh nghia quan he giua B va D} 
%\textcolor{blue}{Note: TODO giai thich quan he B, D}

\subsubsection{Merging overlapping text boxes}
\label{merge:2}
Subsequently, we combine the overlapping text boxes belonging to different $D_i$ results based on the Intersection over Union (IoU). IoU is utilized to determine whether to merge two text boxes into one or separate them into two distinct text boxes. 

Given two bounding boxes, denoted as $B_h$ and $B_k$, the IoU formula is used to determine the overlapped region between them, which is represented by the equation: 
\begin{equation}
IoU_{(B_h, B_k)} = \frac{S(B_h \cap B_k)}{S(B_h \cup B_k)},
\end{equation}
where $S(\cdot)$ is the area of the text box. From empirically experimental findings, we notice that when two text boxes contain different words, IoU of them is always less than 0.5. Therefore, we propose the following formula to ensemble two overlapping text boxes:
\begin{equation}
\begin{cases}
%B_{s} = B_i \cap B_j if IoU >= 0.5 \\
%B_n = B_i - B_i \cap B_j if IoU >= 0.5\\
%B_m = B_j - B_i \cap B_j if IoU >= 0.5
\begin{aligned}
B_{s} &= B_h \cap B_k &\text{if } IoU \geq 0.5, \\
B_n &= B_h - (B_h \cap B_k) &\text{if } IoU < 0.5, \\
B_m &= B_k - (B_h \cap B_k) &\text{if } IoU < 0.5,
\end{aligned}
\end{cases}
\end{equation}
where $B_n$, $B_m$, and $B_s$ are new text boxes after merging two text boxes $B_h$ and $B_k$.

%\begin{figure}[t]
%    \centering
%    \includegraphics[width=3.5in]{im1525.jpg}
%    \caption{before ensemble}
%    \label{before_esm}
%\end{figure}
%\begin{figure}
%    \centering
%    \includegraphics[width=3.5in]{im1525_es.jpg}
%    \caption{after ensemble}
%   \label{after_esm}
%\end{figure}

%\subsection{Stacking Model}

 %\begin{center} -----IN PROGRESS----- \end{center}
\section{EXPERIMENTS}
\label{sec:experiemnts}
\subsection{Implementation Details}

All experiments were performed on Google Colab with the following configuration GPU T4 and 13GB of RAM. %We executed all models using the PaddleOCR library \footnote{\url{https://github.com/PaddlePaddle/PaddleOCR.git}}. 
Detection methods were pre-trained on the ICDAR 2015 \cite{ICDAR2015}  for $1,000$ epochs. The initialized learning rate was $7\times 10^{-3}$ and followed a decay learning rate schedule with a decay factor of $0.9$. After that, the models were fine-tuned for $30$ epochs on VinText dataset \cite{nguyen2021dictionary}, and weight decay regularization was applied with a value of $10^{-4}$. The dataset underwent decoding in BGR format, and various data augmentation techniques were applied, such as horizontal flipping, affine transformations (rotations between $-10^{\circ}$ to $10^{\circ}$), and resizing between $0.5$ and $3$ times the original size.

Recognition methods were pre-trained on MJSynth \cite{MJSynth1, MJSynth2} and SynthText \cite{SynthText} datasets. After that, they underwent a total of 20 epochs of fine-tuning on the VinText dataset \cite{nguyen2021dictionary}. We utilized the Adam optimizer \cite{kingma2014adam} 
with a $\beta_1$ value of 0.9 and $\beta_2$ value of $0.99$. Gradient clipping with a norm threshold of $20.0$ was applied to prevent exploding gradients. The learning rate followed a piecewise schedule, decaying at epoch 20 tow $10^{-4}$ and $10^{-5}$ for subsequent steps. $L_2$ regularization with a factor of 0 was employed.

\subsection{Experimental Settings}

% \subsection{Dataset}

The VinText dataset \cite{nguyen2021dictionary}, a recently proposed Vietnamese text dataset, was used to evaluate methods. This benchmark dataset has a total of 56K text instances and 2,000 images, which consists of 1,200 training images, 500 testing images, and 300 unseen test images. %Some examples of the dataset are shown in Fig. \ref{fig:sampledataset}. 
Most text instances have usually too many patterns and chaotic scenes with scene text instances of various types, appearances, sizes, and orientations. 
%\highlight{what does "busy" mean?}
%\textcolor{red}{da cau nay em mo ta cac text nho? trong hinh}
%\highlight{trong comment o benchmark co ghi su dung unseen image?}. 
We evaluated methods on unseen test images. 

% \begin{figure}[ht]
% \begin{subfigure}{.5\textwidth}
%   \centering
%   % include first image
%   \includegraphics[width=.8\linewidth]{Samples-of-Scene-Text-from-ICDAR-2015.pdf}  
%   \caption{Sample from \textbf{ICDAR 2015} datasets}
%   \label{fig:sub-first}
% \end{subfigure}
% \newline

% \begin{subfigure}{.5\textwidth}
%   \centering
%   % include second image
%   \includegraphics[width=.8\linewidth]{Sample-of-VinText.pdf}  
%   \caption{Sample from \textbf{VinText} datasets}
%   \label{fig:sub-second}
% \end{subfigure}
% \caption{Sample images from the datasets used}
% \label{fig:examplefig}
% \end{figure}

% \begin{figure}[t]
%     \vspace{1mm}
%     \centering
%     \includegraphics[width=\linewidth]{Figures/Sample-of-VinText.pdf}
%     \caption{Sample images from VinText dataset \cite{nguyen2021dictionary}.}
%     \label{fig:sampledataset}
% \end{figure}

% \subsection{Evaluation metric} 

%\highlight{tui em tu sua phan nay: xoa cac cho in dam}

In order to evaluate the performance and effectiveness of methods, we used Character Accuracy ($CA$) metric for recognition task, Precision ($P$), Recall ($R$), and F-measure ($F1$)  metrics for detection task.

\subsection{Vietnamese Scene Text Spotting Benchmark}

% The table presents percentage values for all metrics, and all experiments were conducted using unseen test images from the VinText dataset \cite{nguyen2021dictionary}. Hmean is the primary metric used for comparing models.

\begin{table}[t!]
\renewcommand{\arraystretch}{0.9} 

 \caption{Result of detection methods on the VinText dataset~\cite{nguyen2021dictionary}. $1^{st}$ and $2^{nd}$ places are shown in \textcolor{blue}{\textbf{blue}} and \textcolor{red}{\textbf{red}}, respectively. 
 %\highlight{thu tu cua 2 backbone cua DB bi nguoc so voi cac method khac. Sap xep lai thu tu cho thong nhat voi nhau.}
 }
\label{table:det}

\centering
\resizebox{\columnwidth}{!}{%
\begin{tabular}{c|c|c|c|c|c|c}
\hline
 \textbf{Method} & \textbf{Backbone} & \begin{tabular}[c]{@{}c@{}}\textbf{Pre-trained}\\\textbf{Model}\end{tabular}   & \begin{tabular}[c]{@{}c@{}}\textbf{Fine-tuned on}\\\textbf{VinText~\cite{nguyen2021dictionary}}\end{tabular}  & \textbf{Precision} & \textbf{Recall} & \textbf{Hmean} \\
\hline

\multirow{4}{*}{\centering SAST~\cite{SAST}} & \multirow{4}{*}{\centering ResNet50\_vd~\cite{Resnet50_vd}}  &  \multirow{2}{*}{\centering Total-Text~\cite{Totaltext}} & $\checkmark$ & 87.82 & 53.40 &  66.40     \\
 &   &   &  & 86.09 & 59.98 &  70.70     \\ \cline{3-4}
 &   & \multirow{2}{*}{\centering ICDAR2015~\cite{ICDAR2015}}  & $\checkmark$ & \textcolor{red}{\textbf{89.53}} &	70.56 &	\textcolor{red}{\textbf{78.92}}     \\
 &   &   & & 87.45 &	56.43 &	68.59     \\
 \hline
 
\multirow{2}{*}{\centering DB++~\cite{DB++}}  & \multirow{2}{*}{\centering ResNet50~\cite{Resnet50}} & \multirow{2}{*}{\centering ICDAR2015~\cite{ICDAR2015}} & $\checkmark$ & \textcolor{blue}{\textbf{92.17}}  &   \textcolor{red}{\textbf{71.17}}     &   \textcolor{blue}{\textbf{80.32}}    \\
 &   &  &  & 89.05	&65.18&	75.26    \\
\hline

\multirow{4}{*}{\centering DB~\cite{DB}} &   \multirow{2}{*}{\centering 
 ResNet50\_vd}~\cite{Resnet50_vd} %MobileNetV3~\cite{MobileNetV3}}  
&  \multirow{4}{*}{\centering ICDAR2015~\cite{ICDAR2015}} 
& $\checkmark$ & 85.56 & 65.17    & 73.97 \\
 &     &   & & 84.36 & 56.43  &67.62   \\\cline{2-2} \cline{4-4}
% & $\checkmark$ & 78.73 & 65.34    & 71.41 \\
%  &     &   & & 76.71 &	50.44 &	60.86 \\ \cline{2-2} \cline{4-4}
 &   \multirow{2}{*}{\centering MobileNetV3~\cite{MobileNetV3}}
 &   & $\checkmark$ & 78.73 & 65.34    & 71.41 \\
  &     &   & & 76.71 &	50.44 &	60.86 \\
 \hline

 \multirow{4}{*}{\centering EAST~\cite{EAST}} & \multirow{2}{*}{\centering ResNet50\_vd~\cite{Resnet50_vd}}    & \multirow{4}{*}{\centering ICDAR2015~\cite{ICDAR2015}} & $\checkmark$ & 67.00 & \textcolor{blue}{\textbf{71.86}}    &  69.35     \\
 &     &   &  & 62.59&	54.73&	58.40     \\ \cline{2-2} \cline{4-4}
 &   \multirow{2}{*}{\centering MobileNetV3~\cite{MobileNetV3}}  &   & $\checkmark$ & 68.13 & 69.50    & 68.80   \\ 
 &     &   & & 63.80 &	51.73 &	57.13   \\ 
\hline

\end{tabular}
}
\vspace{-3mm}
\end{table}

\subsubsection{Scene Text Detection}
%We initially conduct experiments using various models on the Vintext dataset, and the obtained results are presented in Table \ref{table:det}. Among the evaluated models, the DB++ model with the ResNet50 backbone achieved the highest precision score of 92.51\% when we pre-train on the ICDAR2015 dataset. Nonetheless, it displayed relatively lower recall and Hmean values, indicating a trade-off between accuracy and comprehensiveness. Conversely, the EAST model, incorporating both ResNet50\_vd and MobileNetV3 backbones, demonstrated moderate performance by striking a balance between precision and recall. The SAST model with the ResNet50\_vd backbone exhibited competitive precision when we pre-train on the Total-Text dataset, although its recall was comparatively lower. However, regarding the ICDAR2015 dataset, the SAST model demonstrated a balanced performance in terms of precision and recall. Furthermore, the DB model, utilizing both MobileNetV3 and ResNet50\_vd backbones, achieved high precision and recall, surpassing other methods in terms of the Hmean metric.

We conducted experiments on the Vintext dataset \cite{nguyen2021dictionary} using various detection methods, including %\highlight{ghi ten cac method va ref vao}
SAST \cite{SAST}, DB++ \cite{DB++}, DB \cite{DB}, EAST \cite{EAST}, in different settings. %\highlight{detection, khong phai recognition}

Table \ref{table:det} showcases the results with DB++ \cite{DB++} outperforming other methods. Specifically, DB++ with the ResNet50 \cite{Resnet50} backbone achieved the highest performance (80.32\% in terms of Hmean) when pre-trained on the ICDAR2015 dataset \cite{ICDAR2015} and fine-tuned on the VinText dataset. On the other hand, SAST \cite{SAST} with the ResNet50\_vd backbone, pre-trained on the ICDAR2015 dataset, demonstrated competitive results across all metrics. However, it is noteworthy that there exists a trade-off between the accuracy and comprehensiveness of these methods, as reflected by relatively lower recall values. In contrast, EAST \cite{EAST} exhibited a balanced performance in terms of precision and recall. These findings shed light on the strengths and limitations of these methods, facilitating informed decision-making when selecting an appropriate scene text detection method for our ensemble learning framework.

% The DB\cite{DB} model with MobileNetV3\cite{MobileNetV3} and ResNet50\_vd \cite{Resnet50_vd} backbones achieved high precision and recall, outperforming other methods in terms of the Hmean metric.

\textbf{Backbone evaluation: } Our investigation focuses on two scene text detection methods, DB\cite{DB} and EAST \cite{EAST}, where we explore the influence of different backbone architectures, ResNet50\_vd \cite{Resnet50_vd} and MobileNetV3 \cite{MobileNetV3}. The comparative results in Table \ref{table:det} highlight the superiority of ResNet50\_vd over MobileNetV3 in terms of performance. These findings emphasize the critical importance of selecting an appropriate backbone architecture for each model, as it significantly influences the effectiveness of text detection and recognition tasks.

\textbf{Effectiveness of fine-tuning models: } Table \ref{table:det} shows that the fine-tuned models on the Vintext dataset \cite{nguyen2021dictionary} exhibit a substantial improvement in results compared to their pre-training counterparts. These findings demonstrate the potency of leveraging the Vintext dataset for fine-tuning to significantly enhance the performance of Vietnamese scene text detection models.

\subsubsection{Scene Text Recognition}

\begin{table}[t!]
\vspace{1mm}
\caption{Result of recognition methods on the VinText \cite{nguyen2021dictionary}. $1^{st}$ and $2^{nd}$ places are shown in \textcolor{blue}{\textbf{blue}} and \textcolor{red}{\textbf{red}}, respectively. We remark that methods were pre-trained on MJSynth \cite{MJSynth1, MJSynth2} and SynthText \cite{SynthText} datasets in default.}
\label{table:rec}

\renewcommand{\arraystretch}{0.9} 
\centering
\resizebox{\columnwidth}{!}{%
\begin{tabular}{c|c|c|c|c}
\hline
 \textbf{Method} & \textbf{Backbone} & \begin{tabular}[c]{@{}c@{}}\textbf{Pre-trained}\\\textbf{Model}\end{tabular}   & \begin{tabular}[c]{@{}c@{}}\textbf{Fine-tuned on}\\\textbf{VinText~\cite{nguyen2021dictionary}}\end{tabular}  & \textbf{Accuracy} \\
\hline
\multirow{3}{*}{\centering SRN \cite{SRN}}  & \multirow{3}{*}{\centering ResNet50\_vd\_fpn \cite{Resnet50_vd}} & $\checkmark$ & $\checkmark$ &   70.73   \\
 &   &  $\checkmark$ &  &  25.81    \\
 & & & $\checkmark$&26.01 \\
 
\hline
\multirow{3}{*}{\centering ABINet \cite{ABINet}}  & \multirow{3}{*}{\centering ResNet45 \cite{resnet45} } &  &  $\checkmark$ &   73.03    \\
 &   &   $\checkmark$ &  & 28.57     \\
 & & & $\checkmark$& 36.89  \\
 \hline
 \multirow{2}{*}{\centering SPIN \cite{SPIN}}  & \multirow{2}{*}{\centering ResNet32 \cite{resnet32}} &  $\checkmark$ &  $\checkmark$ &   \textcolor{red}{\textbf{78.35}}    \\
 &   &   $\checkmark$ &  & 20.69    \\
 \hline

  \multirow{3}{*}{\centering RobustScanner \cite{RobustScanner}}  & \multirow{3}{*}{\centering ResNet31\cite{resnet32}} &  $\checkmark$ &  $\checkmark$ &   \textcolor{blue}{\textbf{80.77} }   \\
 &   &   $\checkmark$ &  & 28.01    \\
 & & &  $\checkmark$& 42.18  \\
 \hline
  \multirow{3}{*}{\centering SAR \cite{SAR}}  & \multirow{3}{*}{\centering ResNet31\cite{resnet32}} &  $\checkmark$ &  $\checkmark$ &   69.30   \\
 &   &   $\checkmark$ &  &  26.10    \\
 & & &  $\checkmark$& 49.21 \\
 \hline
   \multirow{3}{*}{\centering SVTR \cite{SVTR}}  & \multirow{3}{*}{\centering  SVTR\_Tiny \cite{svtr-tiny} } &  $\checkmark$ &  $\checkmark$ &  66.04   \\
 &   & $\checkmark$ &  &   28.04    \\
 & & &  $\checkmark$&  30.04 \\
 \hline
 \multirow{3}{*}{\centering NRTR \cite{NRTR}}  & \multirow{3}{*}{\centering  NRTR\_MTB  \cite{NRTR}} &  $\checkmark$ &  $\checkmark$ &  30.30   \\
 &   &   $\checkmark$ &  &    4.44    \\
 & & &  $\checkmark$ &   28.00 \\
 \hline
  \multirow{3}{*}{\centering RFL \cite{RFL}}  & \multirow{3}{*}{\centering  ResNetRFL  \cite{RFL}} &  $\checkmark$ &  $\checkmark$ &  52.18   \\
 &   &   $\checkmark$ &  &    26.77   \\
 & & &  $\checkmark$&   27.07 \\
 \hline
\end{tabular}
}
\vspace{-3mm}
\end{table}

\begin{figure*}[t]
\vspace{1mm}
\centering
\begin{subfigure}{0.33\textwidth}
  \centering
  \includegraphics[width=\linewidth]{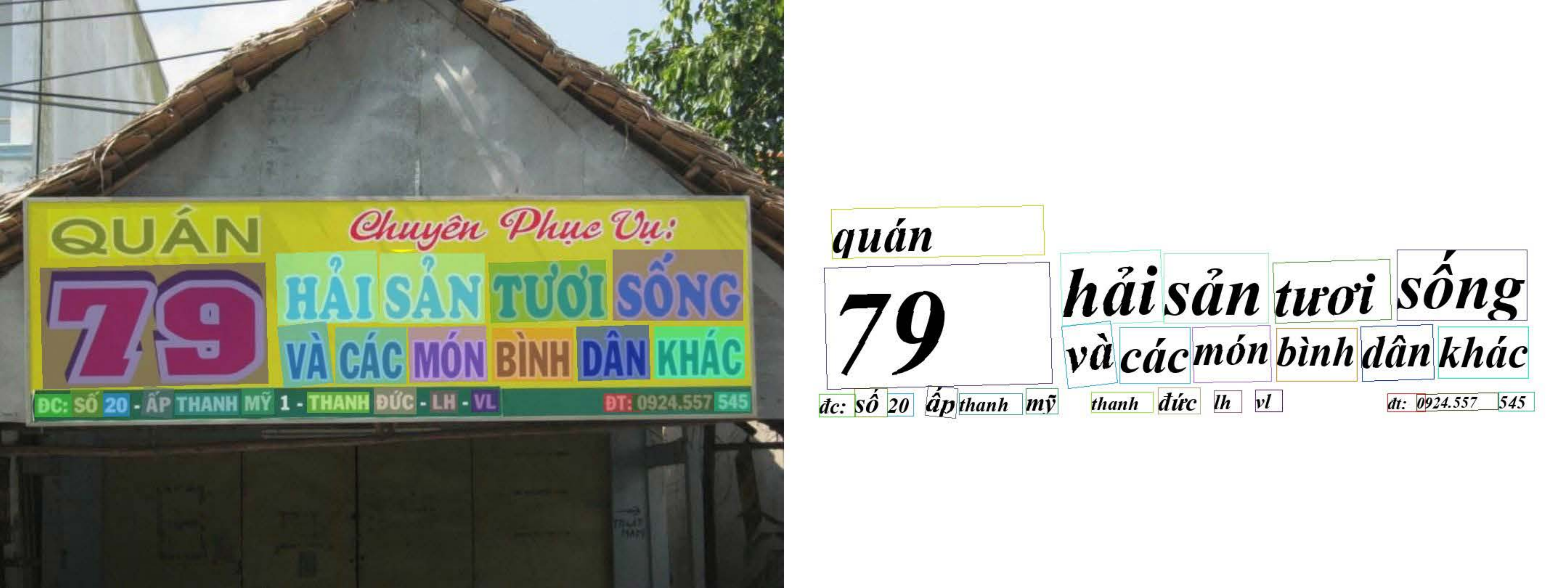}  
  %\caption{DB++~\cite{DB++} and SPIN~\cite{SPIN}}
  \label{fig:model1}
\end{subfigure}%
\hfill
\begin{subfigure}{0.33\textwidth}
  \centering
  \includegraphics[width=\linewidth]{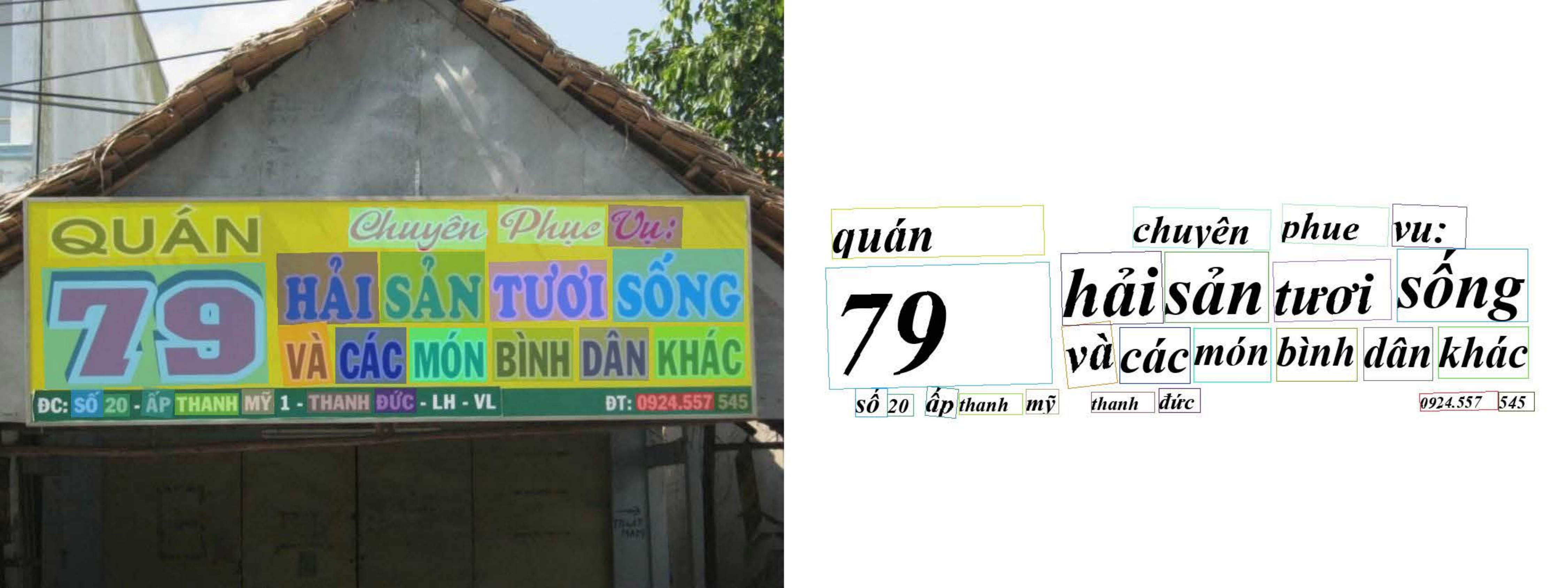}  
  %\caption{SAST~\cite{SAST} and ABINet~\cite{ABINet}}
  \label{fig:model2}
\end{subfigure}%
\hfill
\begin{subfigure}{0.33\textwidth}
  \centering
  \includegraphics[width=\linewidth]{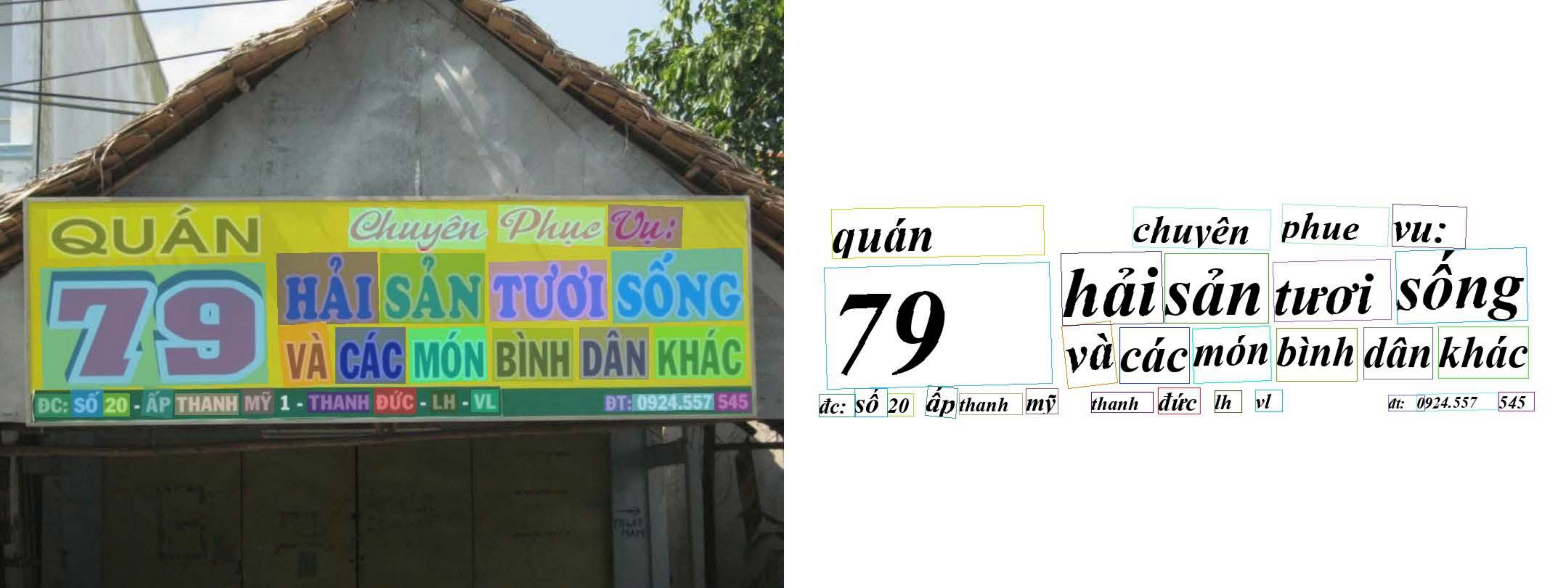}  
  %\caption{Our ensemble method}
  \label{fig:modelensemble}
\end{subfigure}

%% sua tu day

\hfill
\begin{subfigure}{0.33\textwidth}
  \centering
  \includegraphics[width=\linewidth]{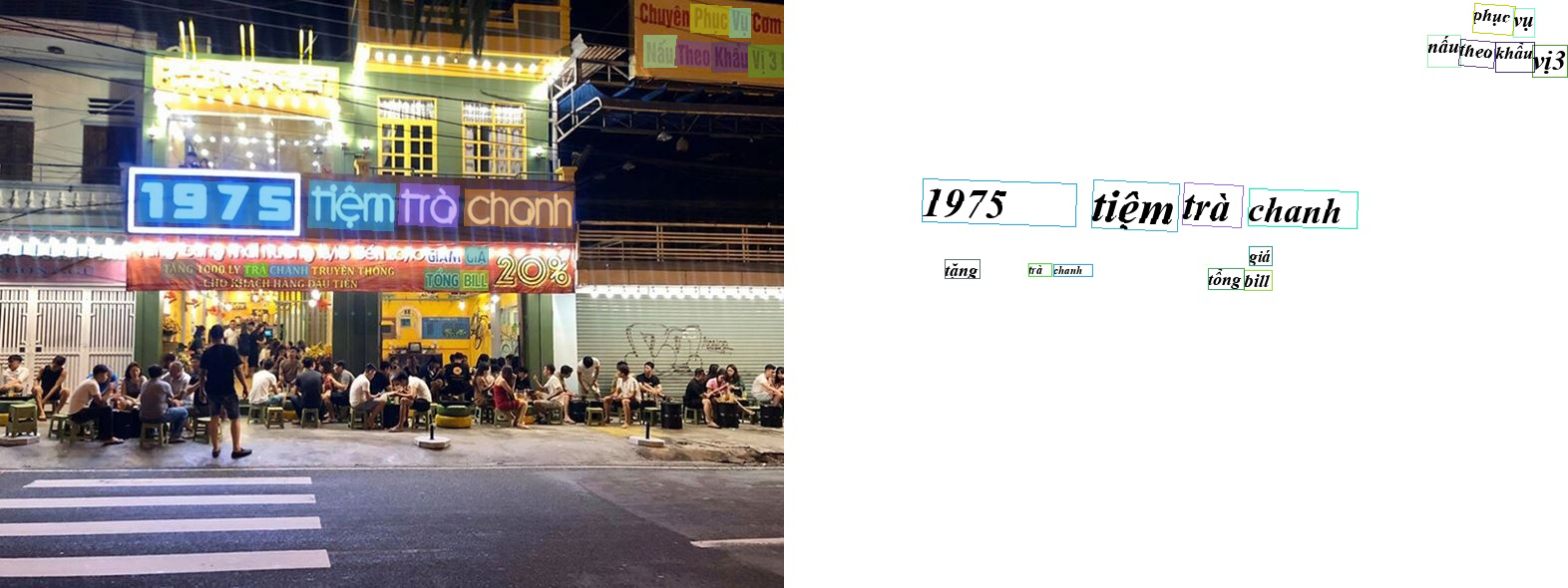}  
  \caption{DB++~\cite{DB++} and SPIN~\cite{SPIN}}
  \label{fig:model11}
\end{subfigure}%
\hfill
\begin{subfigure}{0.33\textwidth}
  \centering
  \includegraphics[width=\linewidth]{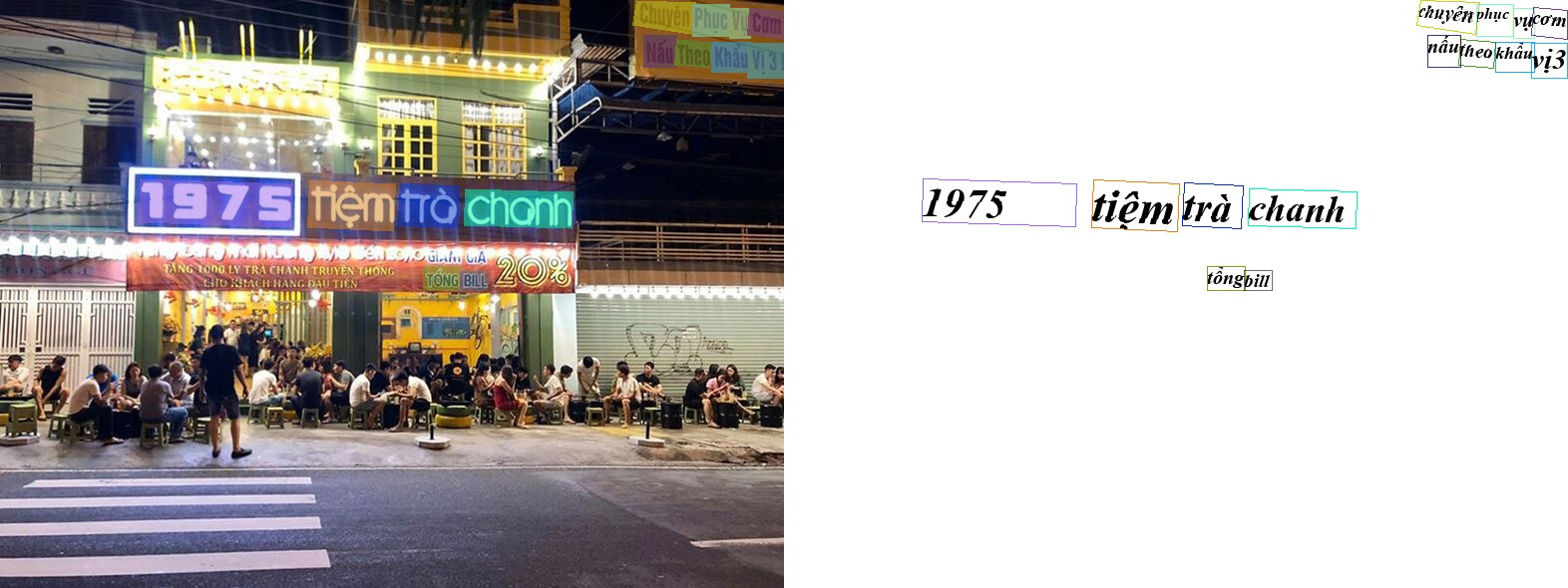}  
  \caption{SAST~\cite{SAST} and ABINet~\cite{ABINet}}
  \label{fig:model22}
\end{subfigure}%
\hfill
\begin{subfigure}{0.33\textwidth}
  \centering
  \includegraphics[width=\linewidth]{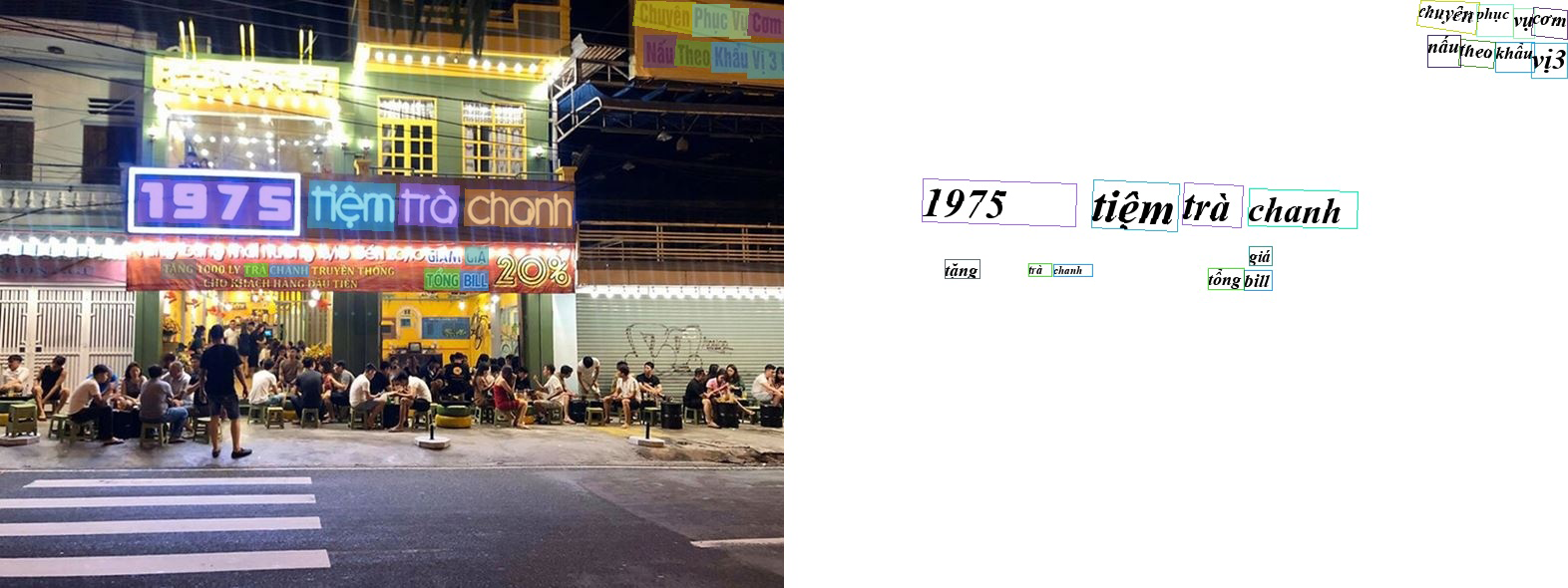}  
  \caption{Our ensemble method}
  \label{fig:modelensemble2}
\end{subfigure}

\caption{Visualization of results of our ensemble learning framework.}
%\highlight{cho them 1 hoac 2 row nua. it qua}
\label{fig:exampleresult}
\vspace{-3mm}
\end{figure*}

We also evaluated the performance of various recognition methods, such as SRN \cite{SRN}, ABINet \cite{ABINet}, SPIN \cite{SPIN}, RobustScanner \cite{RobustScanner}, SAR \cite{SAR}, SVTR \cite{SVTR}, NRTR \cite{NRTR}, RFL \cite{RFL}, in different settings. Table \ref{table:rec} presents a comparative analysis of these methods based on their average accuracy. RobustScanner\cite{RobustScanner}, SPIN\cite{SPIN}, ABINet\cite{ABINet}, and SRN\cite{SRN} stand out with impressive results. RobustScanner achieved the highest accuracy of 80.77\%, followed by SPIN with 78.35\%, ABINet with 73.03\%, and SRN with 70.73\%. These methods consistently demonstrate strong performance. On the other hand, RFL\cite{RFL} and NRTR\cite{NRTR} exhibit relatively lower accuracies of 52.18\% and 30.30\%, respectively.

The experimental results also reveal compelling evidence of the importance of pre-training models on MJSynth \cite{MJSynth1, MJSynth2} and SynthText \cite{SynthText} datasets. Training from scratch on the VinText dataset\cite{nguyen2021dictionary} yields notably inferior performance across all methods. In addition, the results of fine-tuned models exhibit a remarkable improvement, which indicates the necessity of fine-tuning pre-trained models on the VinText dataset\cite{nguyen2021dictionary}. 

\subsection{Ensemble Learning Evaluation}

%sửa lại đoạn này cái, chọn từng model đơn lẻ để thực hiện 2 task detection & recognition mà, nói combine vậy ngta tưởng đoạn đó m ensemble rồi.

%Với cả ensemble là danh từ chứ có phải động từ đâu, dùng động từ combine hoặc fuse gì đó. Chứ bên trên t sửa mấy chỗ r, bên dưới lại dùng ensemble làm động từ nữa .__.

% The models in Table \ref{table:det} and Table \ref{table:rec} are chosen for the detection and recognition stages, respectively, in the Scene Text Spotting problem, we conduct thorough ablation studies using the results presented in Table \ref{table:det} and Table \ref{table:rec}, and for the end-to-end approach we choose PGNet \cite{PGNet}. The objective is to identify models that excel in distinct aspects and can complement each other when merged. 

\begin{table}[t!]
\renewcommand{\arraystretch}{1.0} 
\caption{End-to-end scene text spotting results on the VinText dataset\cite{nguyen2021dictionary}.}
\label{table:e2e}
\centering
% \resizebox{\columnwidth}{!}{%
\begin{tabular}{cl|c|c}
\hline
& \textbf{Method} & \textbf{Char\_acc} & \textbf{F-measure} \\ 
\hline
(1) & DB++~\cite{DB++} and SPIN~\cite{SPIN} & 54.08 & 60.19 \\
(2) & SAST~\cite{SAST} and ABINet~\cite{ABINet} & 52.59 & 58.59 \\
(3) & DB~\cite{DB} and SRN~\cite{SRN} & 50.51 & 56.87 \\
(4) & PGNet (4) \cite{PGNet} & 53.73 & 60.03 \\
\hline
& \textbf{Ensemble learning (Ours):} &  & \\
& (1), (2) & \textbf{58.75} & \textbf{65.21} \\
& (1), (3) & 51.09 & 55.06 \\
& (1), (4) & 56.24 & 64.37 \\
& (1), (2), (3) & 49.83 & 57.33 \\
& (1), (2), (3), (4) & 52.91 & 55.42 \\
\hline
\end{tabular}
% }
\vspace{-5mm}
\end{table}

We carefully analyzed the results of detection and recognition methods in Table \ref{table:det} and Table \ref{table:rec} to select appropriate models for our ensemble learning framework. The objective is to identify models that excel in distinct aspects and can synergistically complement each other when combined. 

In the evaluation of detection methods from Table \ref{table:det}, we meticulously analyze their performance on both the ICDAR2015 \cite{ICDAR2015} and Total-Text datasets \cite{Totaltext}. Notably, DB++\cite{DB++}, EAST\cite{EAST}, and SAST \cite{SAST} stand out in different metrics. DB++ \cite{DB++} showcases high precision, while EAST \cite{EAST} demonstrates good recall, and SAST exhibits a strong F-measure specifically on the Total-Text dataset \cite{Totaltext}.

Shifting our focus to the recognition methods from Table \ref{table:rec}, we assess their average accuracy. SPIN \cite{SPIN}, ABINet \cite{ABINet}, and RobustScanner \cite{RobustScanner} stand out as top performers, demonstrating remarkable accuracy in text recognition. However, upon merging the RobustScanner \cite{RobustScanner} with the detection methods, we observe challenges with overlapping text boxes, leading to a decline in overall performance. As a result, we make the decision to exclude the RobustScanner \cite{RobustScanner} from our final ensemble, striking a balance between detection and recognition performance.

After the individual evaluations, we selected DB++ \cite{DB++}, EAST \cite{EAST}, SAST \cite{SAST})for the detection task, SPIN \cite{SPIN}, ABINet \cite{ABINet}, SRN \cite{SRN} for the recognition task. Additionally, PSNet \cite{PGNet} was chosen as an end-to-end model capable of addressing both detection and recognition aspects. Subsequently, we employed our ensemble algorithm to combine model pairs, harnessing the unique strengths of each end-to-end model. The results of this comprehensive ensemble are presented in Table \ref{table:e2e}, showcasing a powerful and complementary fusion of models for Vietnamese scene text spotting.

While potential of ensemble learning is vast, it is crucial to recognize that mere combination of more models does not always guarantee improved results. In fact, some models, when merged, may yield even worse performance compared to their individual counterparts (see Table \ref{table:e2e}). Our rigorous experimentation has led us to discover that the most optimal ensemble result is achieved by combining pairs of methods DB++\cite{DB++} / SPIN \cite{SPIN} and SAST \cite{EAST}/ ABINet\cite{ABINet}. This carefully curated selection of effective ensemble pairs sheds light on the delicate interplay between methods and underscores the significance of thoughtful model combination for achieving remarkable performance gains in scene text spotting. Indeed, our ensemble framework delivers improved performance in all metrics (See Fig. \ref{fig:exampleresult}).

\section{CONCLUSION}
\label{sec:conclu}
In this paper, we have presented an ensemble learning framework for Vietnamese scene text spotting in urban environments. The proposed technique demonstrates the effectiveness of combining the strengths of different models. Through meticulous experimentation, we identify models with promising performance, strategically designing an ensemble framework that maximizes individual model strengths. Nonetheless, our method does exhibit certain limitations, notably an increase in computational complexity due to the integration of multiple models and instances where certain model combinations yield suboptimal results. In our future research, we aim to address these challenges by focusing on enhancing spelling accuracy during word recognition and reducing the computational complexity of the ensemble model.

\section*{Acknowledgement}

This research is supported by research funding from Faculty of Information Technology, University of Science, Vietnam National University - Ho Chi Minh City.

%\highlight{rut gon paper cho vua du 6 trang}

%\highlight{RUT GON REFERENCES: XAI TEN VIET TAT CUA CONF./JOUR. ArXiv thi phai di kem ID luon. Mot so ref khong co conf./jour. can sua lai}

%\section*{ACKNOWLEDGMENT}

%We express our heartfelt appreciation to our supervisor, Dr. Trung-Nghia Le, for his invaluable guidance and mentorship. We are also deeply grateful to Prof. Minh-Triet Tran and Prof. Quan Vu Hai for their guidance, motivation, and valuable contributions during the review process. Their expertise and insights have greatly enriched the quality of this paper.

%Furthermore, we would like to acknowledge the contributions of our colleagues, friends, and the funding organizations that have provided support and assistance throughout this research journey. Without their involvement and commitment, the completion of this paper would not have been possible. We are sincerely grateful for their contributions and dedication to advancing knowledge in our field.

%In conclusion, we wish to express our profound gratitude to all those who have contributed to this research paper. Their invaluable contributions have been instrumental in its successful completion.

%\bibliographystyle{plain} % We choose the "plain" reference style

 %References
\bibliographystyle{ieeetr}
\bibliography{References}
% \vspace{0pt}
\end{document}